\documentclass[preprint,5p,12pt,twocolumn]{article}

\usepackage{amssymb}
\usepackage[position=top]{subfig}
\usepackage{algorithm,algorithmic}
\usepackage{graphicx}
\usepackage{xspace}
\usepackage{multirow}
\usepackage{rotating}
 \usepackage{url}
\usepackage{gensymb}
\usepackage [ table ]{ xcolor }
\usepackage[top=2cm, bottom=2.2cm, left=1.5cm, right=1.5cm]{geometry}
\usepackage{setspace}
\usepackage{morefloats}
\usepackage[keeplastbox]{flushend}

\usepackage{multirow}
\usepackage{caption}
\usepackage{url}
\usepackage[hyphenbreaks]{breakurl}
\usepackage{amsmath}

\usepackage[affil-it]{authblk}
\usepackage[hidelinks]{hyperref}
\hypersetup{
	colorlinks = true,
	citecolor = blue,
}
\usepackage{pdfpages}
\usepackage{eso-pic}
\definecolor{backcream}{HTML}{FFF0C8}

\AddToShipoutPictureFG*{
	\AtPageUpperLeft{%
		\raisebox{-50pt}{\makebox[\paperwidth]{\noindent\fcolorbox{red}{backcream}{\begin{minipage}{18.5cm}\centering
						{\color{red}\textbf{This manuscript is accepted in Pattern Recognition.} \textbf{Please cite it as:}}\\
						Mozafari, M., Ganjtabesh, M., Nowzari-Dalini, A., Thorpe, S. J., \& Masquelier, T. (2019). Bio-Inspired Digit Recognition Using Reward-Modulated Spike-Timing-Dependent Plasticity in Deep Convolutional Networks. Pattern Recognition. (\url{https://doi.org/10.1016/j.patcog.2019.05.015}).
		\end{minipage}}}}%
	}
}

\begin{document}
\title{\LARGE \textbf{Bio-inspired digit recognition using reward-modulated spike-timing-dependent plasticity in deep convolutional networks}}

\author{Milad~Mozafari$ ^{1}$}
\author{Mohammad~Ganjtabesh$ ^{1,}$\footnote{Corresponding author. \\ \hspace*{0.5cm} Email addresses:\\ \hspace*{1cm} milad.mozafari@ut.ac.ir (MM), \\ \hspace*{1cm} mgtabesh@ut.ac.ir (MG) \\ \hspace*{1cm}  nowzari@ut.ac.ir (AND) \\ \hspace*{1cm}  simon.thorpe@cnrs.fr (SJT) \\ \hspace*{1cm} timothee.masquelier@cnrs.fr (TM).}}
\author{Abbas~Nowzari-Dalini$ ^{1}$}
\author{Simon~J.~Thorpe$ ^{2}$}
\author{Timoth\'ee~Masquelier$ ^{2}$}

\affil{\footnotesize $ ^{1} $ Department of Computer Science, School of Mathematics, Statistics, and Computer Science, University of Tehran, Tehran, Iran}
\affil{\footnotesize $ ^{2} $ CerCo UMR 5549, CNRS -- Universit\'e Toulouse 3, France}

\date{}

\maketitle
\begin{abstract}
The primate visual system has inspired the development of deep artificial neural networks, which have revolutionized the computer vision domain. Yet these networks are much less energy-efficient than their biological counterparts, and they are typically trained with backpropagation, which is extremely data-hungry. To address these limitations, we used a deep convolutional spiking neural network (DCSNN) and a latency-coding scheme. We trained it using a combination of spike-timing-dependent plasticity (STDP) for the lower layers and reward-modulated STDP (R-STDP) for the higher ones. In short, with R-STDP a correct (resp. incorrect) decision leads to STDP (resp. anti-STDP). This approach led to an accuracy of $97.2\%$ on MNIST, without requiring an external classifier. In addition, we demonstrated that R-STDP extracts features that are diagnostic for the task at hand, and discards the other ones, whereas STDP extracts any feature that repeats. Finally, our approach is biologically plausible, hardware friendly, and energy-efficient.
	
	\vspace{0.3cm}
\end{abstract}

\section{Introduction}
\label{intro}
In recent years, deep convolutional neural networks (DCNNs) have revolutionized machine vision and can now outperform human vision in many object recognition tasks with natural images~\cite{rawat2017deep,gu2018recent}. Despite their outstanding levels of performance, the search for brain inspired computational models continues and is attracting more and more researchers from around the world. Pursuing this line of research, a large number of models with enhanced bio-plausibility based on spiking neural networks (SNNs) have emerged. However, SNNs are not yet competitive with DCNNs in terms of recognition accuracy. If DCNNs work well, what is the reason for this increased interest in neurobiological inspiration and SNNs?

To begin with, energy consumption is of great importance. Thanks to the millions of years of optimisation by evolution, the human brain consumes about $20$ Watts~\cite{mink1981ratio} -- roughly the power consumption of an average laptop. Although we are far from understanding the secrets of this remarkable efficiency, the use of spike-based processing has already helped neuromorphic researchers to design energy-efficient microchips~\cite{furber2016large, davies2018loihi}.

Furthermore, employing embedded and real-time systems for artificial intelligence (AI) is important with the advent of small and portable computing devices. In recent years, specialized real-time chips for DCNNs have been released that are capable of fast simulation of pre-trained networks. However, online on-chip training of DCNNs with exact error backpropagation is not yet practical, due to the high-precision and time-consuming operations. In contrast, biologically inspired learning rules such as spike-timing-dependent plasticity (STDP)~\cite{gerstner1996neuronal, bi1998synaptic} can be hardware-friendly, and appropriate for online on-chip training~\cite{yousefzadeh2017hardware}.

These reasons, together with the natural ability of SNNs to handle spatio-temporal patterns, have led researchers to try a range of methods for applying SNNs to visual tasks. The use of hierarchically structured neural networks is a common approach, yet configuring other parameters such as the number of layers, neuron models, information encoding, and learning rules is the subject of much debate. There are shallow~\cite{masquelier2007unsupervised, yu2013rapid} and deep~\cite{lee2016training, o2016deep} SNNs with various types of connectivity structures, such as recurrent~\cite{thiele2017wake}, convolutional~\cite{masquelier2007unsupervised, cao2015spiking, tavanaei2016bio}, and fully connected~\cite{diehl2015unsupervised}. Information encoding is the other aspect of this debate, where rate-based~\cite{merolla2011digital,hussain2014improved,oconnor2014realtime} and temporal coding~\cite{masquelier2007unsupervised, yu2013rapid,diehl2015unsupervised,mostafa2017supervised} are two of the main options. Different learning techniques are also applied to SNNs, from backpropagation~\cite{cao2015spiking,diehl2015fast,wu2017spatio,liu2017mtspike}, tempotron~\cite{yu2013rapid,zhao2015feedforward}, and other supervised techniques~\cite{hussain2014improved,oconnor2014realtime,mostafa2017supervised,ponulak2010supervised,neftci2012event}, to unsupervised STDP and its variants~\cite{diehl2015unsupervised, tavanaei2016acquisition}.

Regarding the pursuit of brain inspiration, STDP-based SNNs are the most biologically plausible ones. Using STDP, the network can successfully extract frequently occurring visual features. However, an unsupervised learning rule alone is not sufficient for decision-making, where external classifiers such as support vector machines (SVMs) and radial basis functions (RBFs), or supervised variants of STDP, are usually required. Digit recognition is one of the well-known pattern recognition problems which has became a standard task to examine the newly proposed methods~\cite{kulkarni2018spiking,hochuli2018handwritten,chang2018deep}. There are several STDP-based SNNs that have been applied to MNIST dataset for digit recognition. For example, Brader et. al.~\cite{brader2007learning} with $96.5\%$, Querlioz et. al.~\cite{querlioz2013immunity} with $93.5\%$, and Diehl and Cook~\cite{diehl2015unsupervised} with $95\%$ of recognition accuracies are successful models the use shallow structure in digit recognition. With a deeper structure, Beyer et. al.~\cite{beyeler2013categorization} achieved a not so good performance of $91.6\%$, however, Kheradpisheh et. al.~\cite{KHERADPISHEH201856} trained a deep convolutional SNN (DCSNN) and increased the accuracy to $98.4\%$.

Researchers have started exploring the potential of using reinforcement learning (RL) in SNNs and DCNNs~\cite{dayan2002reward, daw2006computational, niv2009reinforcement, lee2012neural,steinberg2013causal, schultz2015neuronal,mnih2015human, silver2016mastering}. By RL, the learner is encouraged to repeat rewarding behaviors and avoid those leading to punishments~\cite{sutton1998introduction}. Using supervised learning, the network learns at most what the supervisor knows, while with RL, it is able to explore the environment and learn novel skills (unknown to any supervisor) that increase the fitness and reward acquisition~\cite{silver2017mastering}.

We previously developed a shallow SNN with a single trainable layer~\cite{mozafari2018first}, where the plasticity was governed by reward-modulated STDP (R-STDP). R-STDP is a reinforcement learning rule inspired by the roles of neuromodulators such as Dopamine (DA) and Acetylcholine (ACh) in modulation of STDP~\cite{fremaux2016neuromodulated, brzosko2017sequential}. Our network made decisions about categories of objects solely based on the earliest spike in the last layer without using any external classifier. Together with rank-order encoding and at most one spike per neuron, our network was biologically plausible, fast, energy-efficient, and hardware-friendly with acceptable performance on natural images. However, its shallow structure made it inappropriate for large and complex datasets with high degrees of variations.

In this research, we designed a 3-layer DCSNN with a structure adopted from~\cite{KHERADPISHEH201856}, mainly for digit recognition. The proposed network does not need any external classifier and uses a neuron-based decision-making layer trained with R-STDP. First, the input image is convolved with difference of Gaussian (DoG) filters at various scales. Then, by an intensity-to-latency encoding~\cite{gautrais1998rate}, a spike wave is generated and propagated to the next layer. After passing through multiple convolutional and pooling layers with neurons that are allowed to fire at most once, the spike wave reaches the last layer, where there are decision-making neurons that are pre-assigned to each digit. For each input image, the neuron in the last layer with the earliest spike time or maximum potential indicates the decision of the network.

We evaluated our DCSNN on the MNIST dataset for digit recognition. First, we applied R-STDP only to the last trainable layer and STDP to the first two, achieving $97.2\%$ of recognition performance. Next, we investigated if applying R-STDP to the penultimate trainable layer is helpful. We found that when there are frequent distractors in the input and limitations on the computational resources, using R-STDP instead of STDP in the penultimate layer is beneficial.

The rest of this paper is organized as follows. A precise description of the components of the proposed network, synaptic plasticity, and decision-making is provided in Section $2$. Then, in Sections $3$ and $4$, the performance of the network in digit recognition and the results of applying R-STDP to multiple layers are presented. Finally, in Section $5$, the proposed network is discussed from different points of view and the possible future works are highlighted.

\section{Methods}
\label{matmet}
The focus of this paper is to train and tune a DCSNN by a combination of biologically plausible learning rules, i.e. STDP and R-STDP, for digit recognition. To this end, we modify the network proposed in~\cite{KHERADPISHEH201856} by dropping its external classifier (SVM) and adding a neuron-based decision-making layer, as well as using R-STDP for synaptic plasticity in the higher layers.

The original implementation of the proposed network has done in C++/CUDA and the source code would be available upon request. We have provided a Python\footnote{\url{https://www.python.org/}} script which is a re-implementation of this work using our PyTorch\footnote{\url{https://pytorch.org/}}-based framework called SpykeTorch (available on GitHub\footnote{\url{https://github.com/miladmozafari/SpykeTorch}})~\cite{mozafari2019spyketorch}.

\subsection{Overall Structure}
\label{overall_structure}
\begin{figure*}[!t]
	\begin{center}
		\includegraphics[width = 13cm]{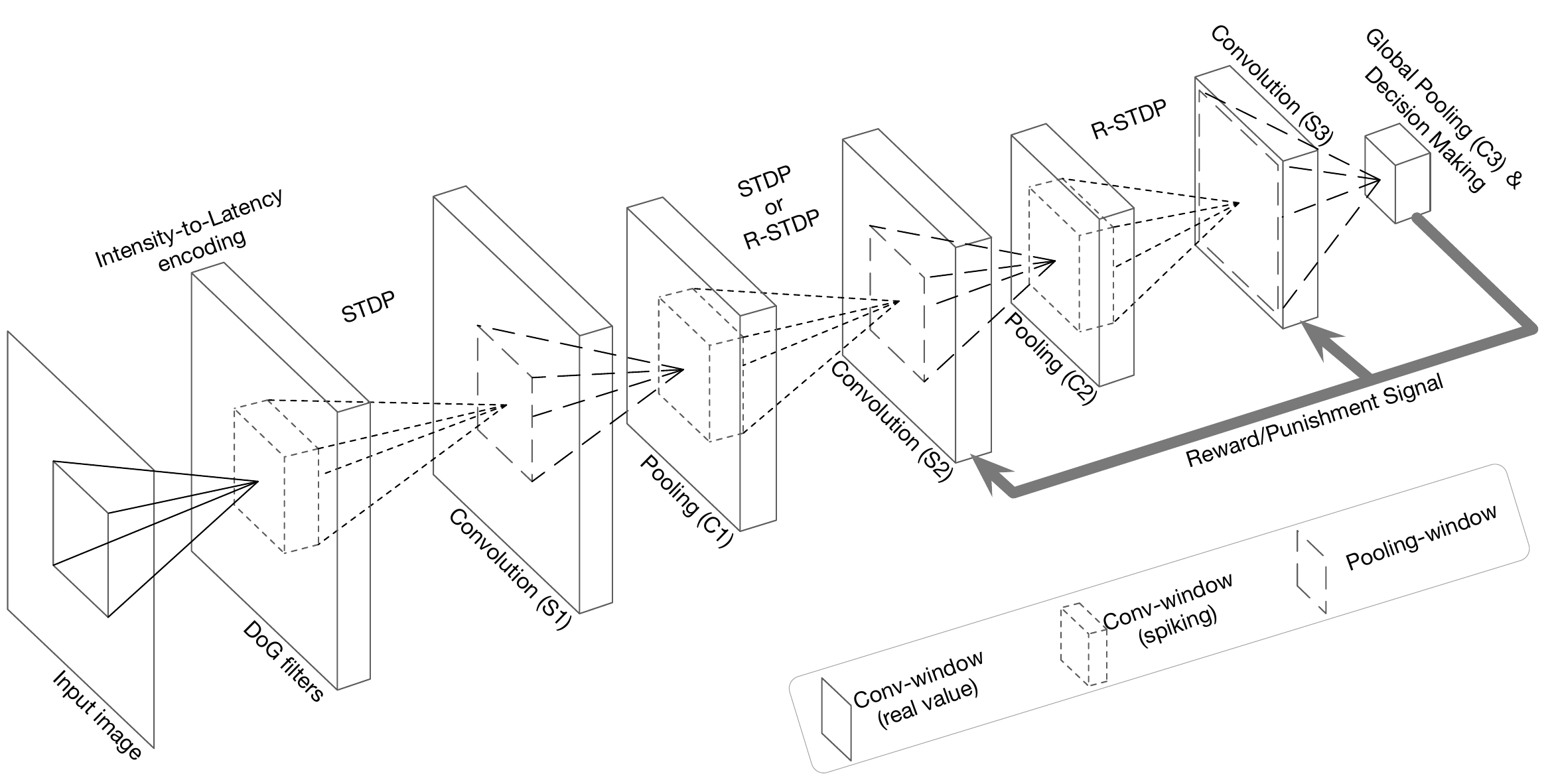}
		\caption{Overall structure of the proposed DCSNN for digit recognition. The input image is convolved by six DoG filters (on- and off-center in three different scales) and converted into spike latencies based on intensity-to-latency encoding scheme. Generated spikes are processed through three spiking convolution-then-pooling layers ($S1-C1$,$S2-C2$, and $S3-C3$). All of the spiking convolution layers are trainable, employing either STDP or R-STDP learning rules. The network makes its decision in $C3$, where neurons are pre-assigned to digits, and the decision is the digit assigned to the neuron with either the maximum internal potential or the earliest spike time. Regarding the decision, a reward/punishment (reinforcement) signal will be generated globally, which is received by R-STDP-enabled layers.}
		\label{fig_model}
	\end{center}
\end{figure*}

The proposed network has six layers, that are three convolutional layers $(S1, S2, S3)$, each followed by a pooling layer $(C1, C2, C3)$. To convert MNIST images into spike waves, they are convolved with six DoG filters and encoded into spike times by the intensity-to-latency encoding scheme. Plasticity of the afferents of convolutional layers is done by either STDP or R-STDP learning rules. The ultimate layer ($C3$) is a global pooling in which neurons are pre-assigned to each digit. These neurons are indicators for the network's decision. Figure~\ref{fig_model} plots the outline of the proposed network.

Details of the input spike generation, functionality of each layer, learning rules, and decision-making process are given in the rest of this section.

\subsection{Input Spike Waves}
\label{inp_spk_trn}
On each input image, On- and Off-center DoG filters of three different scales are applied with zero padding. Window sizes are set to $3\times 3$, $7\times 7$, and $13\times 13$, where their standard deviations $(\sigma_1, \sigma_2)$ are $(3/9, 6/9)$, $(7/9, 14/9)$, and $(13/9, 26/9)$, respectively. We keep the ratio of $2$ between two sigmas in each scale. These values are chosen practically based on the statistics of input images.

From the output of the DoG filters, all the values below $50$ are ignored and the remaining values are descendingly sorted, denoting the order of spike propagation.

In order to speed up running time of our parallel implementation of the network, we distribute the ordered input spikes equally into a fixed number of bins. All the spikes in each bin are propagated simultaneously.

\subsection{Convolutional Layers}
\label{convolution}
Each convolutional layer ($S$-layer) in the proposed network contains several 2-dimensional grids of Integrate-and-Fire (IF) neurons, that constitute the feature maps. Each neuron in a $S$-layer has a fixed 3-dimensional input window of afferents with equal width and height, and a depth equal to the number of feature maps in the previous layer. The firing threshold is also set to be the same across all the neurons in a single layer.

In each time step, the internal potential of each IF neuron is increased by the incoming spikes within its input window using the magnitude of the corresponding synaptic weights. These neurons have no leakage and if a neuron reaches the firing threshold, it will emit a single spike, after which remains silent until the next input image is fed to the network. For each new input, the internal potentials of all neurons are reset to zero. Note that a weight-sharing mechanism is applied for all neurons in a single feature map.

\subsection{Pooling Layers}
\label{pooling}
Pooling layers ($C$-layers) in our network are employed for introducing position invariance and reducing information redundancy. Each $C$-layer has the same number of feature maps as its previous $S$-layer and there is a one-to-one association between maps of the two layers.

There are two types of pooling layers in our network: spike-based and potential-based. Both types have a 2-dimensional input window and a particular stride. Each neuron in the spike- and potential-based $C$-layers, indicates the earliest spike time and the maximum potential of the neurons within its input window, respectively.

\subsection{Decision-Making and Reinforcement Signal}
\label{decision}
As mentioned before, neurons in the final $C$-layer ($C3$) perform global pooling on their corresponding $S3$ grids. These neurons are labeled with input categories such that each $C3$ neuron is assigned to single category, but each category may be assigned to multiple neurons. Using the labeled neurons in $C3$, the decision of the network for each input is the label of the neuron with the earliest spike time or the maximum internal potential depending on the type of pooling layer.

When the decision of the network has been made, it will be compared with the original label of the input image. If they match (or mismatch), a reward (or punishment) signal will be generated globally. This signal will be received by layers that employ R-STDP rule for synaptic plasticity (see the next section).

\subsection{Plasticity}
\label{plastic}
STDP is an unsupervised learning rule which functions based on the ordering of pre- and post-synaptic spikes~\cite{gerstner1996neuronal, bi1998synaptic}. For each pair of pre- and post-synaptic spikes, if the ordering is pre-then-post (or post-then-pre), then the synapse will be potentiated (or depressed). STDP is useful to learn repetitive spike patterns among a large set of incoming spike trains ~\cite{masquelier2008spike, gilson2011stdp, brette2012computing, masquelier2017stdp, masquelier2018optimal}.

R-STDP is a reinforcement learning rule, in which the STDP rule takes the role of neuromodulators into account~\cite{fremaux2016neuromodulated}. Environmental feedback can be encoded into the release of neurotransmitters. Unlike STDP, which is blind to these feedbacks, R-STDP modifies the polarity of weight changes based on the released neuromodulators. For instance, the release of DA can be considered as a rewarding feedback that encourages the neuron's behavior. Conversely, the release of ACh can signal punishment and result in discouraging the neuron's behavior~\cite{brzosko2017sequential}. For differences between STDP and R-STDP in action, we refer the readers to~\cite{mozafari2018first}.

Here, we use either STDP or R-STDP to train each of the trainable layers. We define a general formula that can be employed for both rules as follows:
\footnotesize
\begin{align}
\delta_{ij}&=
\begin{cases}
\alpha \phi_r^{} a_r^+ + \beta \phi_p^{} a_p^- & \hspace{0.07cm} \mbox{if} \ t_j - t_i \leq 0,\vspace{0.4cm}\\
\alpha \phi_r^{} a_r^- + \beta \phi_p^{} a_p^+ &
\begin{aligned}
& \mbox{if} \ t_j - t_i > 0,\\ & \mbox{or neuron} \ j \ \mbox{never fires}
\end{aligned} 
\end{cases}\\
\Delta w_{ij}&=\delta_{ij}(w_{ij})(1-w_{ij}),
\end{align}
\normalsize
where $i$ and $j$ refer to the post- and pre-synaptic neurons, respectively, $\Delta w_{ij}$ is the amount of weight change for the synapse connecting the two neurons, and $a_r^+$, $a_r^-$, $a_p^+$, and $a_p^-$ scale the magnitude of weight change. Furthermore, to specify the direction of weight change, we set $a^+_r, a^+_p \ge 0$ and $a^-_r, a^-_p \le 0$. Our learning rule only needs the sign of the spike time difference, not the exact value, and uses an infinite time window. The other parameters, say $\phi_r^{}$, $\phi_p^{}$, $\alpha$, and $\beta$, are employed as controlling factors which will be set differently for each learning rule. Moreover, we initially set the synaptic weights with random values drawn from a normal distribution with mean $0.8$ and standard deviation $0.02$~\cite{KHERADPISHEH201856}.

To apply STDP to a $S$-layer, the controlling factors are $\phi_r^{} = 1$, $\phi_p^{} = 0$, $\alpha = 1$, and $\beta = 0$. For R-STDP, the values of $\alpha$ and $\beta$ depends on the reinforcement signal generated by the decision-making layer. If a ``reward'' signal is generated, then $\alpha = 1$ and $\beta = 0$, whereas if a ``punishment'' signal is generated, then $\alpha = 0$ and $\beta = 1$. The reinforcement signal can be ``neutral'' as well, for which $\alpha = 0$ and $\beta = 0$. The neutral signal is useful when we have unlabeled data. Similar to our previous work~\cite{mozafari2018first}, we apply adjustment factors by $\phi_r^{} = \frac{N_{miss}}{N}$ and $\phi_p^{} = \frac{N_{hit}}{N}$, where $N_{hit}$ and $N_{miss}$ denote the number of samples that are classified correctly and incorrectly over the last batch of $N$ input samples, respectively. Note that the plasticity rules are applied for each image, while $\phi_r^{}$ and $\phi_p^{}$ are updated after each batch of samples.

When implementing plasticity in each trainable layer, we employ a $k$-winner-take-all mechanism, by which only $k$ neurons are eligible for plasticity. These neurons cannot belong to a same feature map because of the weight sharing strategy through the entire map. Choosing the $k$ winners is first based on the earliest spike times, and then by selecting those with the highest internal potentials. During the training process, when a winner is indicated, a $r\times r$ inhibition window, centering the winner neuron, will be applied to all the feature maps, preventing them from being selected as the next winners for the current input image.

Additionally, we multiply the term $w_{ij}(1-w_{ij})$ by $\delta_{ij}$ which not only keeps weights between the range $[0,1]$, but also stabilizes the weight changes as the weights converge.

\section{Task 1: Solving 10-class Digit Recognition}
\label{task_1}
In this task, we train our proposed DCSNN using the $60000$ images of handwritten digits from the MNIST dataset. Then the recognition performance of the network is tested over the $10000$ unseen samples provided by the author of the dataset.

\subsection{Network Configuration}
\label{task_1_netconfig}
The input layer consists of six feature maps corresponding to the six DoG filters (three scales with both on- and off-center polarities), each contains the spike latencies. After that, there are three $S$-layers, each followed by a $C$-layer with specific parameters that are summarized in Tables~\ref{convolution_params}--\ref{learning_params}.

The first two $S$-layers have specific thresholds, however the last one has an ``infinite'' threshold. This means that $S3$ neurons never fire for an input image. Instead, they accumulate all the incoming spikes as their internal potential. When there are no more spikes to accumulate, a manual spike time, greater than the previous spike times, will be set for all of the $S3$ neurons.

$C1$ and $C2$ neurons perform spike-based local pooling, whereas $C3$ neurons apply potential-based global pooling which is consistent with the behavior of $S3$ neurons. Accordingly, the proposed network makes decisions based on the maximum potential among $C3$ neurons. In other words, the label associated to the $C3$ neuron with the maximum potential will be selected as the network's decision for each input image.

It is worth mentioning that in this task, we drop the weight stabilizer term from the plasticity equation for layer $S3$. Instead, we manually limit the weights between $0.2$ and $0.8$. This modification is done based on our experimental results, where we found that it is better to continue synaptic plasticity even in late iterations.

\subsection{Training and Evaluation}
\label{task_1_rslt}
We trained the network in a layer-by-layer manner. Layers $S1$ and $S2$ were trained by STDP for $10^5$ and $2\times 10^5$ iterations, respectively. In each iteration, an image was fed to the network and the learning rates were multiplied by $2$ after each $500$ iterations. We kept doing multiplication  while $a_r^+$ was less than $0.15$.

\begin{figure}[t]
	\begin{center}
		\includegraphics[width = 9cm]{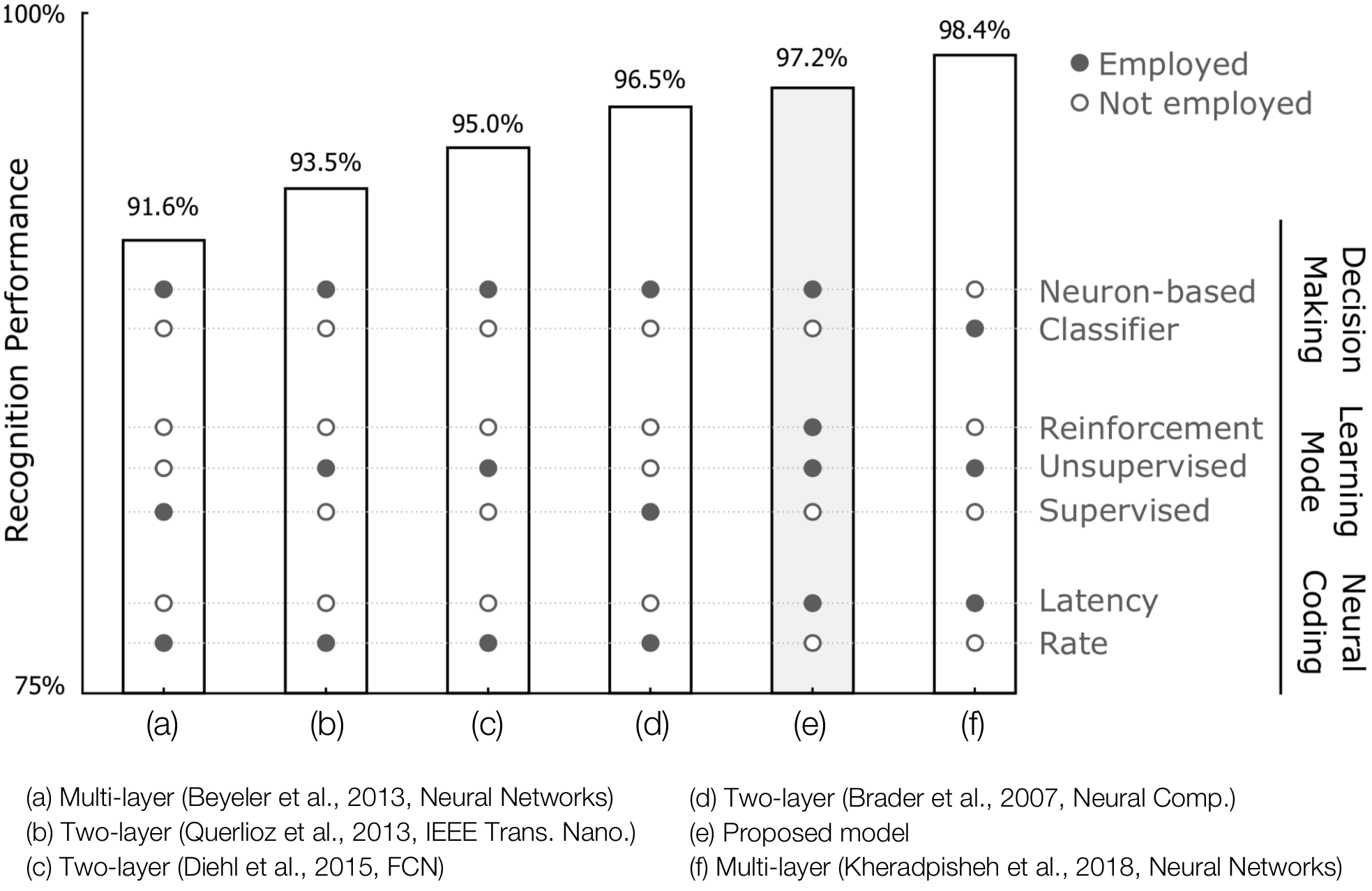}
		\caption{Recognition performance of several available SNNs trained by STDP-based learning rules on the MNIST dataset, as well as the proposed network.}
		\label{fig_compare}
	\end{center}
\end{figure}

Training of $S3$ was governed by R-STDP employing the reinforcement signal produced by layer $C3$. We trained $S3$ for $4\times 10^7$ iterations in total and evaluated the recognition performance after each $6\times 10^4$ iterations (i.e. presenting the whole set of training samples). Our network achieved $97.2\%$ of performance, which is higher than most of the previous STDP-based SNNs evaluated on the MNIST dataset (see Figure~\ref{fig_compare}). A comprehensive comparison of SNNs trained on MNIST dataset with various learning rules is provided in~\cite{tavanaei2018deep}. It is worth mentioning that our network ranked as the second best even though it does not use an external classifier. This increases not only the biological plausibility, but also the hardware friendliness of the network~\cite{KHERADPISHEH201856,mozafari2018first}.

We also examined the application of R-STDP to the penultimate $S$-layer ($S2$) as well, but this did not improve the performance in this particular task. By analyzing the influence of R-STDP on $S2$, we hypothesized that it can help extracting target-specific diagnostic features, if there are frequent distractors among the input images and the available computational resources are limited. Our next task is designed to examine this hypothesis.

\subsection{Comparison to Error Backpropagation}
One of the advantages of using STDP in early layers is that it decreases the chance of overfitting by extracting common features. Unlike error backpropagation which is data-hungry, the proposed network performs well even if we reduce the number of training samples. In order to demonstrate this advantage, we compared the performance of our network to a supervised SNN proposed in~\cite{mostafa2017supervised} while training them with varying numbers of training samples. The chosen supervised SNN uses a similar latency coding scheme with exactly one spike per neuron. Moreover, it achieved a similar performance using the whole training set.

\begin{figure}[t!]
	\begin{center}
		\includegraphics[width = 9cm]{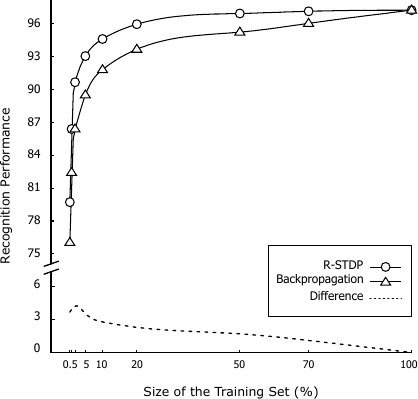}
		\caption{Comparison of recognition performances of the proposed network and a supervised SNN~\cite{mostafa2017supervised} while training them with restricted training set.}
		\label{fig_supcompare}
	\end{center}
\end{figure}

As shown in Figure~\ref{fig_supcompare}, it is clear that our network regained almost all of its generalization power even with $20\%$ of the training samples. Conversely, the performance of the supervised SNN was tightly bound to the number of training samples which means achieving the highest accuracy requires the whole training set. Regarding the fact that humans learn with a few number of samples, the proposed network appears more realistic.

\section{Task 2: R-STDP Can Discard Non-Diagnostic Features}
\label{task_2}
The goal of this task is to distinguish two specific digits while other digits are presented as distractors. This task is conducted over a subset of MNIST images employing a smaller network, which allows us to analyze the advantages of applying R-STDP to more than one layer. This part of the paper is started by introducing a handmade problem and a simple two-layer network to solve it, for the sake of a better illustration. Then, we show the same outcome using images of the MNIST dataset.

\subsection{Handmade Problem}
\begin{figure}[!t]
	\centering
	\par
	\subfloat[\label{fig_toya}]{
		\includegraphics[width = 9cm]{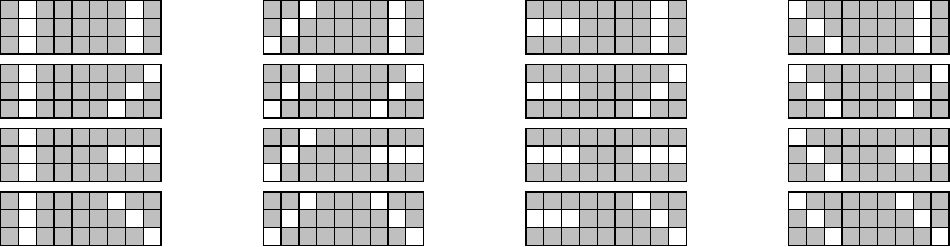}
	}\par
	\subfloat[\label{fig_toyb}]{
		\includegraphics[width = 9cm]{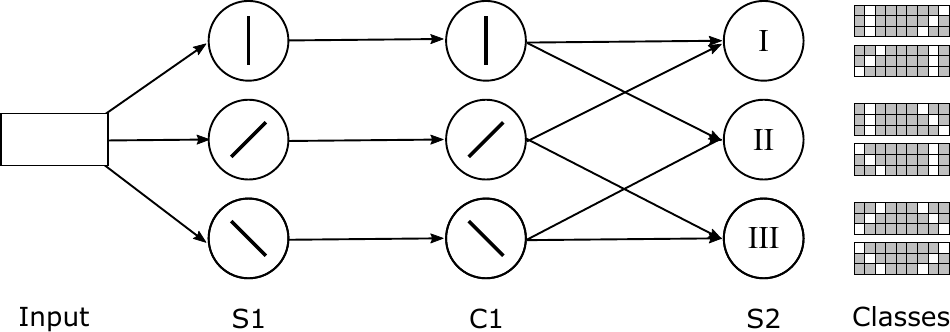}
	}
	\caption{(a) Input set for the handmade problem. It contains all combinations of bars in four different orientations. For each input, spikes are simultaneously propagated from the white tiles. A $3\times 3$ gap is placed between the two oriented bars in order to prevent neurons seeing parts of the two bars at the same time. (b) A simple network with two trainable layers ($S1$ and $S2$) that solves a 3-class recognition task with layer $S2$ as the decision-making layer. Each class is the position invariant combination of two different bars. These classes are defined in a way that the horizontal bar carries no distinguishing information (distractor), while others are vital for final recognition (targets). Since there are only three feature maps in $S1$, the task is fully solvable if and only if the network learns all the three targets in this layer.}
	\label{fig_toy}
\end{figure}

We prepare a set of artificial images of size $9\times 3$, each contains two oriented bars in its left and right most $3\times 3$ regions. As shown in Figure~\ref{fig_toya} , we make all of the possible combinations of bars from four different orientations. We characterize three classes of target inputs by pair-wise combinations of three differently oriented bars (Figure~\ref{fig_toyb}), leaving the other inputs as distractors. All of the inputs are fed to the network in a random order.

We design a minimal two-layer spiking network that distinguishes the three target classes, if and only if it learns the three diagnostic oriented lines by the neurons in the penultimate $S$-layer, i.e. $S1$ (see Figure~\ref{fig_toyb}). According to the target classes, the horizontal bar carries no diagnostic information and learning this bar is a waste of neuronal resources. To this end, we put three feature maps in layer $S1$ whose neurons have $3\times 3$ input windows. After a spike-based global pooling layer, we added three neurons that have $1\times 1\times 3$ input windows.

We tried both STDP and R-STDP learning rules for training $S1$, while the last layer was always trained by R-STDP in order to make decisions. Note that when we used R-STDP on two layers, the plasticity of both layers was postponed until arrival of the reinforcement signal, and thus they were trained simultaneously. Moreover, the reinforcement signal for non-target images was ``neutral'' regardless of the decision of the network.

According to the results, the network with STDP-trained $S1$ failed most of the times to extract the correct orientations. In contrast, training $S1$ with R-STDP enabled the network to solve the problem without any failure. The obtained results agree with the nature of STDP, which extracts frequent features regardless of the outcome. Since the probability of the appearance of each oriented bar is the same, the chance of extracting all the diagnostic bars is $25\%$ for STDP (the probability of discarding the non-diagnostic feature among the four is $1/4$).

We implemented a similar task with the MNIST dataset. In each instance of the task, two digits are selected as the targets and the other digits marked as distractors. The goal is to distinguish the two target digits using a small number of features, while the network receives images of all digits.

\subsection{Network Configuration}
\label{task_2_netconfig}
The number of layers and their arrangement is the same as the network in Task 1, however there are several differences. First, MNIST images are fed to the network without application of DoG filters. Second, we put fewer feature maps with different input window sizes and thresholds in all of the $S$-layers. Third, $S3$ neurons have specific thresholds, that allow them to fire, thus $C3$ performs a spike-based global pooling operation. In addition, for each category, there is only one neuron in $C3$. The values of parameters are summarized in Tables~\ref{convolution_params}--\ref{learning_params}.

\begin{table*}[!t]
	\begin{center}
		\caption{Values of the $S$-layer parameters used in each digit recognition task.}
		\label{convolution_params}
		\begin{tabular}{|c|c|c|c|c|}
			\cline{2-5}
			\multicolumn{1}{c|}{~} & \multirow{2}{*}{Layer} & Number of & Input Window & \multirow{2}{*}{Threshold} \\
			\multicolumn{1}{c|}{~} & ~ & Feature Maps & (Width, Height, Depth) & ~ \\
			\hline
			\parbox[t]{3mm}{\multirow{3}{*}{\rotatebox[origin=c]{90}{Task 1}}} & $S1$ & $30$ & $(5,5,6)$ & $15$ \\
			~ & $S2$ & $250$ & $(3,3,30)$ & $10$ \\
			~ & $S3$ & $200$ & $(5,5,250)$ & $\infty$ \\
			\hline
			\parbox[t]{3mm}{\multirow{3}{*}{\rotatebox[origin=c]{90}{Task 2}}} & $S1$ & $10$ & $(5,5,1)$ & $5$ \\
			~ & $S2$ & $x\in\{2,4,...,20\}\cup\{30,40\}$ & $(15,15,10)$ & $40$ \\
			~ & $S3$ & $2$ & $(1,1,x)$ & $0.9$ \\
			\hline
		\end{tabular}
	\end{center}
\end{table*}

\begin{table*}[!t]
	\begin{center}
		\caption{Values of the $C$-layer parameters used in each digit recognition task.}
		\label{pooling_params}
		\begin{tabular}{|c|c|c|c|c|}
			\cline{2-5}
			\multicolumn{1}{c|}{~} & \multirow{2}{*}{Layer} &  Input Window & \multirow{2}{*}{Stride} & \multirow{2}{*}{Type} \\
			\multicolumn{1}{c|}{~} & ~ & (Width, Height) & ~ & ~ \\
			\hline
			\parbox[t]{3mm}{\multirow{3}{*}{\rotatebox[origin=c]{90}{Task 1}}} & $C1$ & $(2,2)$ & $2$ & Spike-Based \\
			~ & $C2$ & $(3,3)$ & $3$ & Spike-Based  \\
			~ & $C3$ & $(5,5)$ & $0$ & Potential-Based \\
			\hline
			\parbox[t]{3mm}{\multirow{3}{*}{\rotatebox[origin=c]{90}{Task 2}}} & $C1$ & $(2,2)$ & $2$ & Spike-Based \\
			~ & $C2$ & $(14,14)$ & $0$ & Spike-Based  \\
			~ & $C3$ & $(1,1)$ & $0$ & Spike-Based \\
			\hline
		\end{tabular}
	\end{center}
\end{table*}

\begin{table*}[!t]
	\begin{center}
		\caption{Values of parameters for the synaptic plasticity used in each digit recognition task.}
		\label{learning_params}
		\begin{tabular}{|c|c|c|c|c|c|c|c|}
			\cline{2-8}
			\multicolumn{1}{c|}{~} & Layer & $a_r^+$ & $a_r^-$ & $a_p^+$ & $a_p^-$ & $k$ & $r$ \\
			\hline
			\parbox[t]{3mm}{\multirow{3}{*}{\rotatebox[origin=c]{90}{Task 1}}} & $S1$ & $0.004$ & $-0.003$ & $0$ & $0$ & $5$ & $3$ \\
			~ & $S2$ & $0.004$ & $-0.003$ & $0$ & $0$ & $8$ & $2$  \\
			~ & $S3$ & $0.004$ & $-0.003$ & $0.0005$ & $-0.004$ & $1$ & $0$ \\
			\hline
			\parbox[t]{3mm}{\multirow{3}{*}{\rotatebox[origin=c]{90}{Task 2}}} & $S1$ & $0.004$ & $-0.003$ & $0$ & $0$ & $1$ & $0$ \\
			~ & $S2$ & $0.04$ & $-0.03$ & $0.005$ & $0.04$ & $1$ & $0$  \\
			~ & $S3$ & $0.004$ & $-0.003$ & $0.0005$ & $-0.004$ & $1$ & $0$ \\
			\hline
		\end{tabular}
	\end{center}
\end{table*}

\subsection{Training and Evaluation}
\label{task_2_rslt}
We performed the task for all pairs of digits ($\binom{10}{2}=45$ pairs), and different number of feature maps in $S2$, ranging from $2$ to $40$. For each pair, the network was trained over $10^4$ samples from the training set and tested over $100$ testing samples of each target digit ($200$ testing samples in total). $S1$ was separately trained by STDP, while on $S2$ both STDP and R-STDP were examined. Plasticity in $S3$ was always governed by R-STDP.

\begin{figure}[!t]
	\begin{center}
		\includegraphics[width = 9cm]{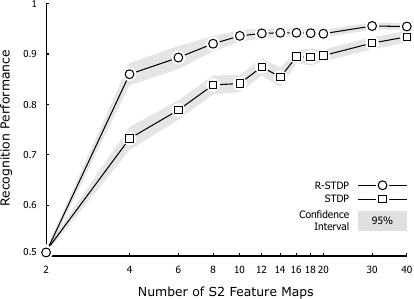}
		\caption{Comparison of recognition performance when STDP or R-STDP is applied to $S2$ in Task 2. Superiority of R-STDP over STDP in finding diagnostic features (and discarding distractors) is clearly shown when there are small number of feature maps in $S2$. As the number of feature maps increases, STDP has more chance to extract features for target digits, therefore approaches the performance of applying R-STDP. With respect to the number of feature maps, R-STDP shows a monotonic behavior in increasing the performance and confidence level, however, STDP is not as robust as R-STDP, specially in terms of confidence level.}
		\label{fig_distract}
	\end{center}
\end{figure}

\begin{figure*}
	\begin{center}
		\includegraphics[width = 15cm]{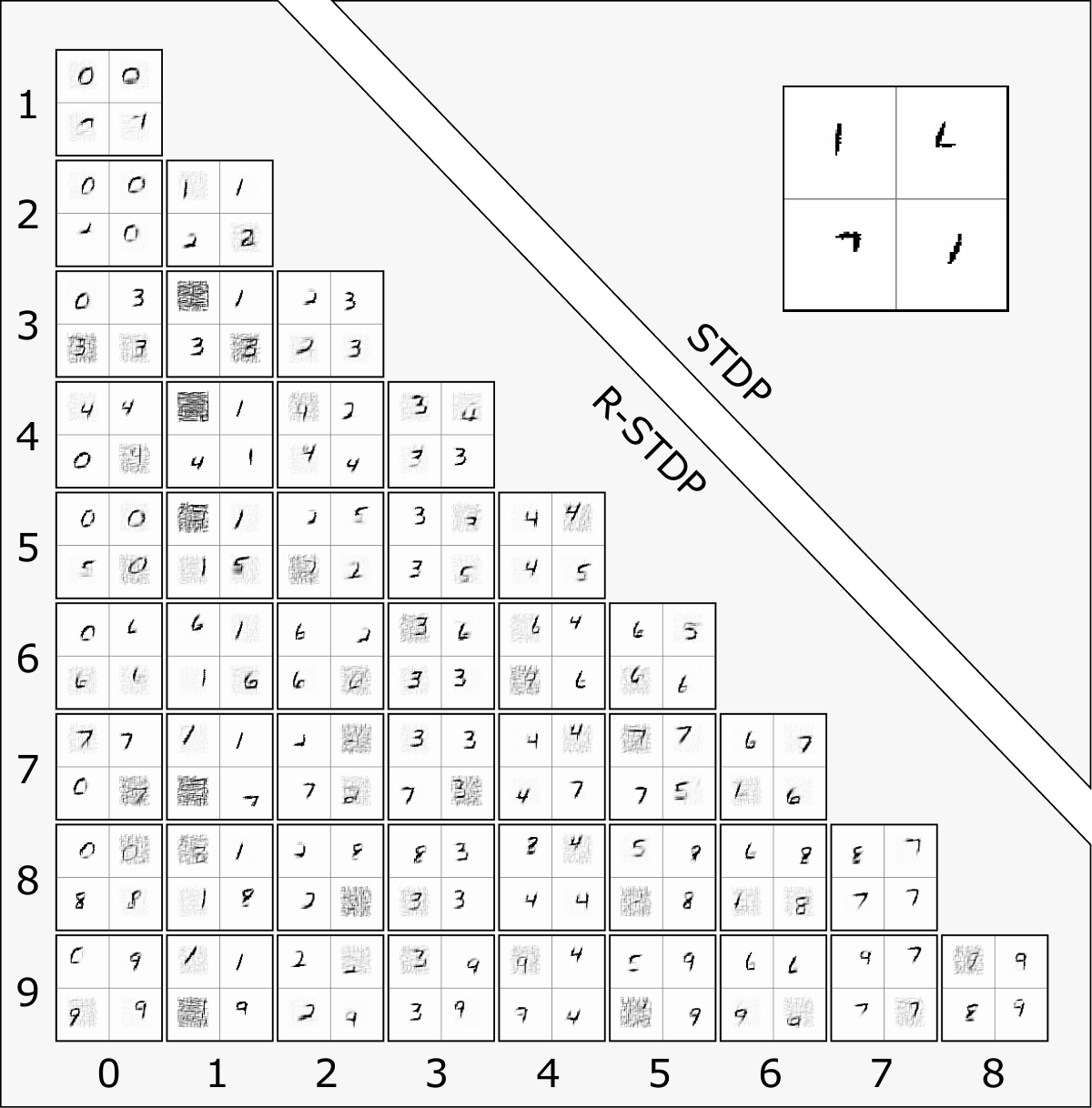}
		\caption{Reconstruction of features extracted by $S2$ neurons in Task 2, when the number of feature maps is limited by four. The lower part of the figure illustrates features that are extracted by applying R-STDP to $S2$ afferents, solving each of the pair-wise digit recognition tasks (corresponding to row and column indices). The existence of non-trained or semi-trained feature maps (dark or gray squares) is due to the fact that other features that are representative enough, win the plasticity competition for all images, leaving no chance for others to learn or converge. In the upper part, the only four features obtained by applying STDP to $S2$ are shown. }
		\label{fig_distract_feat}
	\end{center}
\end{figure*}

For each particular number of feature maps in $S2$, we calculated the average recognition performance of the network for both cases of using STDP and R-STDP over all of the digit pairs. As shown in Figure~\ref{fig_distract}, the results of the experiments clearly support our claim. As we have fewer neuronal resources, using R-STDP is more beneficial than STDP. However, by increasing the number of feature maps in $S2$, STDP also has the chance to extract diagnostic features and fills the performance gap. According to Figure~\ref{fig_distract_feat}, R-STDP helps $S2$ neurons to extract target-aware features. In contrast, using STDP, $S2$ neurons are blind to the targets and extract the same features for all the pairs (the same random seed is used for all of the tasks).

We acknowledge that the network could achieve higher accuracies if its parameters are tuned for each pair of digits. Besides, in order to get the results in a feasible time for large number of simulations, we had to limit the number of iterations and use faster learning rates which again, degrade the performance.

\section{Discussion}
\label{disc}
With the development of DCNNs, AI is now able to solve sophisticated problems such as visual object recognition, in which humans used to be the masters~\cite{rawat2017deep}. Most of the early DCNNs employ supervised learning techniques to be trained for a specific task. However, supervised learning limits the learner to the information provided by the supervisor. Recently, researchers have found that reinforcement learning could be the new game changer and started to combine DCNNs with reinforcement learning techniques, also known as deep reinforcement learning. This combination has introduced networks that are able to achieve super-human skills by finding solutions that have never been tested by a human before~\cite{mnih2015human,silver2016mastering,silver2017mastering}.

Despite the revolutionary performance of DCNNs in machine vision, SNNs have increasingly being used to solve complex vision tasks. SNNs are not yet as impressive as DCNNs, however, they possess a great spatio-temporal capacity by which they are able to fill the performance gap or even bypass DCNNs. Additionally, SNNs have been shown to be energy efficient and hardware friendly, which makes them suitable to be deployed in AI hardware~\cite{furber2016large, davies2018loihi} for online on-chip learning.

In online learning, no training set is provided a priori, or the environment is so dynamic that the prior knowledge would not be enough to solve the task. The goal of an online learner is to improve its performance and adapt to the new conditions while it is experiencing various circumstances or observing different instances in an online manner. There are problems for which using online learning is a must such as, anomaly and fraud detection, analysis of financial markets, and online recommender systems, to mention a few~\cite{hoi2018online}.

Therefore, online learning on hardware (online on-chip learning) needs fast and hardware-friendly algorithms. Because of high-precision and time-consuming operations for supervised error backpropagation, DCNNs are hardly an appropriate choice for this type of learning. However, this has motivated researchers to modify the error backpropagation algorithm in order to make it more hardware-friendly~\cite{neftci2017event}. Conversely, computation and communication with spikes can be extremely fast and applicable in low precision platforms~\cite{rueckauer2017conversion}. Moreover, biologically inspired learning rules such as STDP and R-RSTDP has proved to be quite efficient on real-time hardware~\cite{yousefzadeh2017hardware}.

STDP has been successfully applied to SNNs for digit recognition~\cite{diehl2015unsupervised}, however, because of its unsupervised nature, a supervised readout is required to achieve high accuracies~\cite{KHERADPISHEH201856}. Apart from the fact that supervised readouts (e.g. SVMs) are not biologically supported, they are not appropriate for hardware implementation.

In this paper, we presented a DCSNN for digit recognition which is trained by a combination of STDP and R-STDP learning rules. R-STDP is a biologically plausible reinforcement learning rule~\cite{fremaux2016neuromodulated}, appropriate for hardware implementation as well. This combination is inspired by the brain where the synaptic plasticity is mostly unsupervised, taking into account the local activity of pre- and post-synaptic neurons, but it can be modulated by the release of neuromodulators, reflecting the feedbacks received from the surrounding environment~\cite{gerstner1996neuronal,bi1998synaptic,fremaux2016neuromodulated,brzosko2017sequential}.

The proposed network extracts regularities in the input images by STDP in lower layers and uses R-STDP in higher layers to learn rewarding behavior (accurate decision-making). R-STDP enables us to use a neuron-based decision-making layer instead of complex external classifiers. In this network, information is encoded in the earliest spike time of neurons. The input image is convolved with multiple DoG filters and converted into spike latencies by the intensity-to-latency encoding scheme. Spike propagation as well as plasticity are done in a layer-wise manner except that all the R-STDP-enabled layers are trained simultaneously. When the spike wave reaches the decision-making layer, the label of the neuron with either the maximum potential or the earliest spike time will be selected as the network's decision. Comparing the decision to the original label of the input, a reward/punishment (reinforcement) signal is globally generated which modulates the plasticity in R-STDP-enabled trainable layers.

First, we applied our network to the whole MNIST dataset employing R-STDP only on the last trainable layer. Our network achieved $97.2\%$ of recognition accuracy which is higher than most of the previously proposed STDP-based SNNs. We illustrated that unlike the supervised error backpropagation which is data-hungry, the proposed network achieves almost its ultimate recognition power even with a small portion of the training samples ($\thicksim20\%$). Next we investigated if applying R-STDP to more than one layer is helpful, by allowing the penultimate trainable layer to use R-STDP as well. Reviewing the results of a handful of different experiments revealed that if the input set is polluted with frequent distractors, applying R-STDP to more than one layer helps ignoring distractors and extracting diagnostic features using lower computational resources than a STDP-enabled layer.

The proposed network inherits bio-plausability, energy efficiency and hardware friendliness from its ancestors~\cite{KHERADPISHEH201856,mozafari2018first}, but is now able to solve harder and complex recognition tasks thanks to its deeper structure. Our network has two fundamental features that elevate its energy efficacy in comparison to the other deep networks, particularly DCNNs: (1) Communication with spikes: Using spike/no-spike regime, the accumulation of the incoming spikes can be implemented by energy-efficient ``accumulator" units. In the other, since DCNNs use rate-based encoding, their hardware implementation needs energy-consuming ``multiply-accumulator" units. (2) At most one spike per neuron: SNN hardware can be implemented in an event-driven approach, where spikes are considered as events. Using at most one spike per neuron, while most neurons do not fire at all, results in minimal energy consumption. SNNs with single spike per neuron have got more popularity recently. Apart from the energy-efficiency, they have shown to be competitive if the parameter values are tuned well for the target task~\cite{falez2019multi,vaila2019deep}.

In order to improve the performance of the proposed network, we tested various configurations and values for parameters. Here, we survey our findings regarding the results of these tests:

\textbf{Deciding based on the maximum potential in Task 1.}
We tried decision-making based on the earliest spike in the first step, but the results were not competing. By exploring the miss-classified samples, we found that the digits with common partial features are the most problematic ones. For instance, digits $1$ and $7$ share a skewed vertical line ($1$ is mostly written skewed in MNIST). Let's say a neuron has learned to generate the earliest spike for digit $1$. Obviously, it will generate the earliest spike for many samples of digit $7$ as well, resulting in high rate of false alarms. Thus, we decided to omit the threshold and consider the maximum potential as the decision indicator. Since the network has to wait until the last spike, this approach increased the computation time, however it solved the aforementioned problem.

\textbf{R-STDP is only applied to the last layer in Task 1.}
As we showed in Task 2 and our previous work~\cite{mozafari2018first}, R-STDP can be better than STDP when computational resources are limited. In Task 1, the goal was to increase the performance of the network to an acceptable and competing level. Since MNIST is a dataset with high range of variations for each digit, we had to put enough feature maps in each layer to cover them. As a consequence, the large number of feature maps and the fact that there were no distractors in the inputs, STDP worked well in extracting useful intermediate features.

\textbf{Learning in the ultimate layer is slower than the penultimate one in Task 2.}
In Task 2, $S2$ and $S3$ were trained simultaneously. If $S3$ modifies synaptic weights as fast as $S2$, each change in $S2$ may change a previously good behavior of $S3$ into a bad one, which then ruin $S2$, and this cycle may continue forever, destroying stability of the network.

\section{Conclusion}
In this paper, we proposed a bio-realistic DCSNN trained by the combination of STDP and R-STDP which decides on the category of an input image using neurons in its ultimate layer. Applying STDP to the lower layers helped the network extracting simple frequent features while the application of R-STDP to the higher layers guided the network toward extracting complex diagnostic features as the combinations of early features.

Using three trainable layers and at most one spike per neuron, the proposed network reached $97.2\%$ of recognition performance on MNIST dataset and ranked as the second best among other STDP-based networks. We showed that, unlike the supervised counterpart, our network needs much less training data ($\thicksim20\%$) to reach its ultimate performance. Besides, employing R-STDP on multiple layers helps the network to avoid frequent non-diagnostic features and extract task-specific ones by neurons in the mid-layer as well as the last layer.

We believe that our DCSNN can be further improved from multiple aspects. As mentioned before, potential-based decision-making degrades the computational efficiency. One solution to overcome false alarms, due to common partial features, while having spike-based decision-making is to engage inhibitory neurons. Inhibitory neurons can suppress a neuron if it receives something more than the preferred stimulus (e.g. the horizontal line of digit $7$ would inhibit those neurons responsible for digit $1$, thus producing no false alarm). Another aspect would be performance improvement by designing a population-based decision-making layer. For instance, one can consider a group of earliest spikes instead of one and apply major voting, which may affect the training strategy as well. Another direction for future work would be investigation of the layer-wise application of R-STDP for intermediate layers where the reward/punishment signal can be generated based on other activity measures such as sparsity, information content, and diversity.

\section*{Acknowledgment}
This research received funding from Iran National Science Foundation: INSF (No. 96005286), the European Research Council under the European Union's 7th Framework Program (FP/2007--2013)/ERC Grant Agreement no. 323711 (M4 project), and has been supported by the Center for International Scientific Studies \& Collaboration (CISSC) and French Embassy in Iran. The authors thank NVIDIA GPU Grant Program for supporting computations by providing a high-tech GPU.

\footnotesize
\bibliographystyle{elsarticle-num}

\end{document}